\documentclass{article}
\makeatletter
\IfFileExists{corl_2026.sty}{%
  \usepackage[final]{corl_2026}
}{%
  \usepackage[letterpaper,textwidth=5.5in,textheight=9in,top=1in,headheight=0pt,headsep=0pt]{geometry}
  \usepackage{times}
  \usepackage[numbers,sort&compress]{natbib}
  \renewcommand{\maketitle}{%
    \begin{center}
      {\Large\bfseries \@title \par}\vspace{0.9em}
      {\normalsize \begin{tabular}[t]{c}\@author\end{tabular}\par}
    \end{center}\vspace{0.4em}}
}
\@ifpackageloaded{natbib}{}{\usepackage[numbers,sort&compress]{natbib}}
\makeatother

\usepackage[T1]{fontenc}
\usepackage{amsmath,amssymb,amsthm,mathtools,bm}
\usepackage{booktabs,multirow,makecell}
\usepackage{xcolor}
\usepackage{graphicx}
\usepackage{enumitem}
\usepackage{microtype}
\usepackage{needspace}
\usepackage{flafter}
\usepackage{placeins}
\usepackage{tikz}
\usetikzlibrary{arrows.meta,positioning,calc,shapes.geometric,decorations.pathreplacing,backgrounds,fit}
\usepackage{pgfplots}
\pgfplotsset{compat=1.17}
\usepgfplotslibrary{fillbetween,groupplots}
\usepackage{caption}
\usepackage[colorlinks=true,linkcolor=black,citecolor=black,urlcolor=black]{hyperref}
\usepackage[capitalise,nameinlink]{cleveref}

\theoremstyle{plain}
\newtheorem{theorem}{Theorem}
\newtheorem{proposition}{Proposition}

\theoremstyle{definition}

\theoremstyle{remark}

\crefname{assumption}{Assumption}{Assumptions}
\crefname{problem}{Problem}{Problems}
\newcommand{\runin}[1]{\par\addvspace{2pt}\Needspace{4\baselineskip}\noindent\textbf{#1}\enspace}

\newcommand{\E}{\mathbb{E}}
\renewcommand{\P}{\mathbb{P}}
\newcommand{\F}{\mathcal{F}}

\newcommand{\1}{\mathbb{1}}

\newcommand{\dV}{\Delta V}

\newcommand{\ours}{\textsc{CGEM}}
\newcommand{\asc}{\textsc{ASC}}
\newif\ifshowplaceholders
\showplaceholdersfalse

\newcommand{\key}[1]{\textit{#1}}  

\definecolor{cV}{RGB}{38,84,160}
\definecolor{cPos}{RGB}{34,139,84}
\definecolor{cNeg}{RGB}{200,55,55}
\definecolor{cF}{RGB}{150,150,150}
\definecolor{cBank}{RGB}{232,240,252}
\definecolor{cAlias}{RGB}{245,236,214}
\colorlet{cGray}{black!50}

\pgfplotsset{
  annals/.style={
    axis lines*=left,
    axis line style={line width=0.4pt,black},
    every tick/.style={line width=0.4pt,black},
    tick align=outside, major tick length=2.2pt, minor tick length=1.2pt,
    tick label style={font=\scriptsize},
    label style={font=\footnotesize},
    title style={font=\footnotesize\itshape, yshift=-3pt},
    legend style={draw=none,fill=none,font=\scriptsize,row sep=-1.5pt},
    legend cell align=left,
    every axis plot/.append style={line width=0.55pt,black},
    mark size=1.5pt,
    xtick pos=bottom, ytick pos=left,
    clip=false,
  }
}

\title{Remember What Changes the Odds:\\Critic-Gated Event Memory for Long-Horizon Manipulation}

\author{Xuanzhuo Liu$^{1}$ \quad Chuankang Li$^{2}$ \quad Maoqing Yao$^{2}$ \quad Guanghui Ren$^{2}$\\
{\small $^{1}$Stanford University \quad $^{2}$AgiBot}\\
{\small\texttt{xuanzliu@stanford.edu}, \texttt{\{lichuankang, renguanghui, yaomaoqing\}@agibot.com}}}
\begin{document}
\maketitle

\begin{abstract}
In long-horizon manipulation, a robot may need to take different actions under identical observations, depending on task progress and the outcomes of earlier attempts.
We refer to this ambiguity as \emph{decision aliasing}.
Resolving it requires the policy to maintain memory of its execution history.
To reduce the cost of processing full observation histories, recent methods build compact event memories using perceptual, semantic, or kinematic cues.
However, these criteria may miss events that matter for task success.
We instead propose an outcome-grounded principle: \emph{remember events that substantially change the predicted probability of eventual success}.
Following this principle, we introduce \ours{}, a critic-gated event memory that uses a critic to select events with large estimated changes in success probability.
Positive and negative changes are encoded as progress and failure records, respectively.
Failure records are discarded once the corresponding subtask is completed.
In a ten-subtask diagnostic environment with exactly computed success probabilities, \ours{} achieves $88.0\%$ success, compared with $0.0\%$ for both a 36-step observation window and a kinematic event memory.
Removing failure records reduces success to $1.7\%$.
\end{abstract}

\section{Introduction}
\label{sec:intro}

\begin{figure}[t]
\centering
\begin{tikzpicture}[x=1cm,y=1cm,font=\footnotesize,
  arr/.style={-{Stealth[length=4pt]},line width=.65pt,black!65},
  box/.style={rounded corners=3pt,line width=.65pt},
  card/.style={rounded corners=2pt,minimum height=.35cm,font=\footnotesize,inner sep=3pt},
  home/.pic={
    \fill[black!3,rounded corners=2pt] (-1.8,-.38) rectangle (1.8,.38);
    \draw[black!25,rounded corners=2pt] (-1.8,-.38) rectangle (1.8,.38);
    \draw[black!45] (-1.6,-.24)--(1.6,-.24);
    \fill[black!30] (-1.2,-.24) rectangle (-.8,-.1);
    \fill[cNeg] (-1.08,-.1) rectangle (-.94,-.04);
    \draw[black!70,line width=1.8pt,line cap=round] (-.15,-.24)--(.05,.2)--(.55,.25);
    \draw[black!70] (.55,.25)--(.65,.1) (.55,.25)--(.73,.22);
    \draw[black!45,fill=black!8] (.9,-.24) rectangle (1.45,-.03);
  }]
\node[anchor=west,font=\small\bfseries] at (0,.65) {(a) Same observation, different next actions};
\foreach \x/\label/\col in {2/{Before an attempt}/cV,6.75/{After failure}/cNeg,11.5/{After success}/cPos}{
  \pic at (\x,-.15) {home};
  \node[text=\col] at (\x,.32) {\label};
}
\node at (2,-.8) {No record $\rightarrow$ attempt A};
\node at (6.75,-.8) {Failed A $\rightarrow$ retry B};
\node at (11.5,-.8) {Completed $\rightarrow$ advance};
\draw[black!20] (0,-1.13)--(13.5,-1.13);
\node[anchor=west,font=\small\bfseries] at (0,-1.48) {(b) Proposed memory architecture for $\pi_{0.5}$};
\draw[box,draw=black!35,fill=black!3] (0,-2.2) rectangle (2.4,-4.85);
\node[font=\footnotesize\bfseries] at (1.2,-2.48) {Current inputs};
\node[align=center] at (1.2,-3.25) {RGB\\proprioception\\force/torque};
\node[align=center] at (1.2,-4.3) {Language\\instruction};
\draw[box,draw=orange!70!black,fill=orange!7] (3.05,-2.2) rectangle (6.05,-4.85);
\node[font=\footnotesize\bfseries] at (4.55,-2.48) {Critic + event gate};
\node at (4.55,-2.93) {Success prediction $\hat V$};
\draw[cV,line width=1pt] (3.4,-3.93)--(3.85,-3.93)--(3.85,-3.65)--(4.4,-3.65)--(4.4,-3.35)--(4.95,-3.35)--(4.95,-3.85)--(5.65,-3.85);
\fill[cPos] (4.4,-3.35) circle (1.7pt);
\fill[cNeg] (4.95,-3.85) circle (1.7pt);
\node at (4.55,-4.48) {Write if $|\Delta\hat V|\ge\delta$};
\draw[box,draw=cPos!65!black,fill=cPos!5] (6.75,-2.2) rectangle (9.65,-4.85);
\node[font=\footnotesize\bfseries] at (8.2,-2.48) {Event memory};
\node at (8.2,-2.88) {At most $K$ records};
\node[card,draw=cPos,fill=cPos!10,text=cPos!65!black] at (8.2,-3.35) {Progress + context};
\node[card,draw=cNeg,fill=cNeg!8,text=cNeg!75!black] at (8.2,-3.88) {Failure + strategy};
\node[align=center] at (8.2,-4.52) {Delete obsolete failures};
\draw[box,draw=cV!70!black,fill=cV!5] (10.35,-2.2) rectangle (13.5,-4.85);
\node[font=\footnotesize\bfseries] at (11.925,-2.48) {$\pi_{0.5}$ policy};
\node[draw=cV!60,fill=white,rounded corners=2pt,align=center,minimum width=2.7cm,minimum height=.7cm] (vlm) at (11.925,-3.32) {VLM backbone\\gated memory attention};
\node[draw=cV!60,fill=white,rounded corners=2pt,minimum width=2.7cm,minimum height=.5cm] (act) at (11.925,-4.25) {Action expert};
\draw[arr] (vlm.south)--(act.north);
\draw[arr] (2.4,-3.45)--(3.05,-3.45);
\draw[arr] (6.05,-3.25)--node[above,font=\scriptsize]{write}(6.75,-3.25);
\draw[arr] (9.65,-3.32)--node[above,font=\scriptsize]{read}(10.35,-3.32);
\draw[arr] (8.2,-4.85)--(8.2,-5.2)--node[below,font=\scriptsize]{memory context}(4.55,-5.2)--(4.55,-4.85);
\draw[arr] (1.2,-2.2)--(1.2,-1.9)--node[above,font=\scriptsize]{current observation + instruction}(11.925,-1.9)--(11.925,-2.2);
\draw[arr] (13.5,-4.25)--(13.75,-4.25)--(13.75,-5.75)--node[below,font=\scriptsize]{execute action chunk and observe the outcome}(1.2,-5.75)--(1.2,-4.85);
\end{tikzpicture}
\caption{\textbf{Decision aliasing and the proposed \ours{} architecture.} (a)~Memory distinguishes untried, failed, and completed interactions under the same observation. (b)~A critic selects events for a bounded memory of progress and failure records. Both the critic and policy read the memory.}
\label{fig:system}
\end{figure}
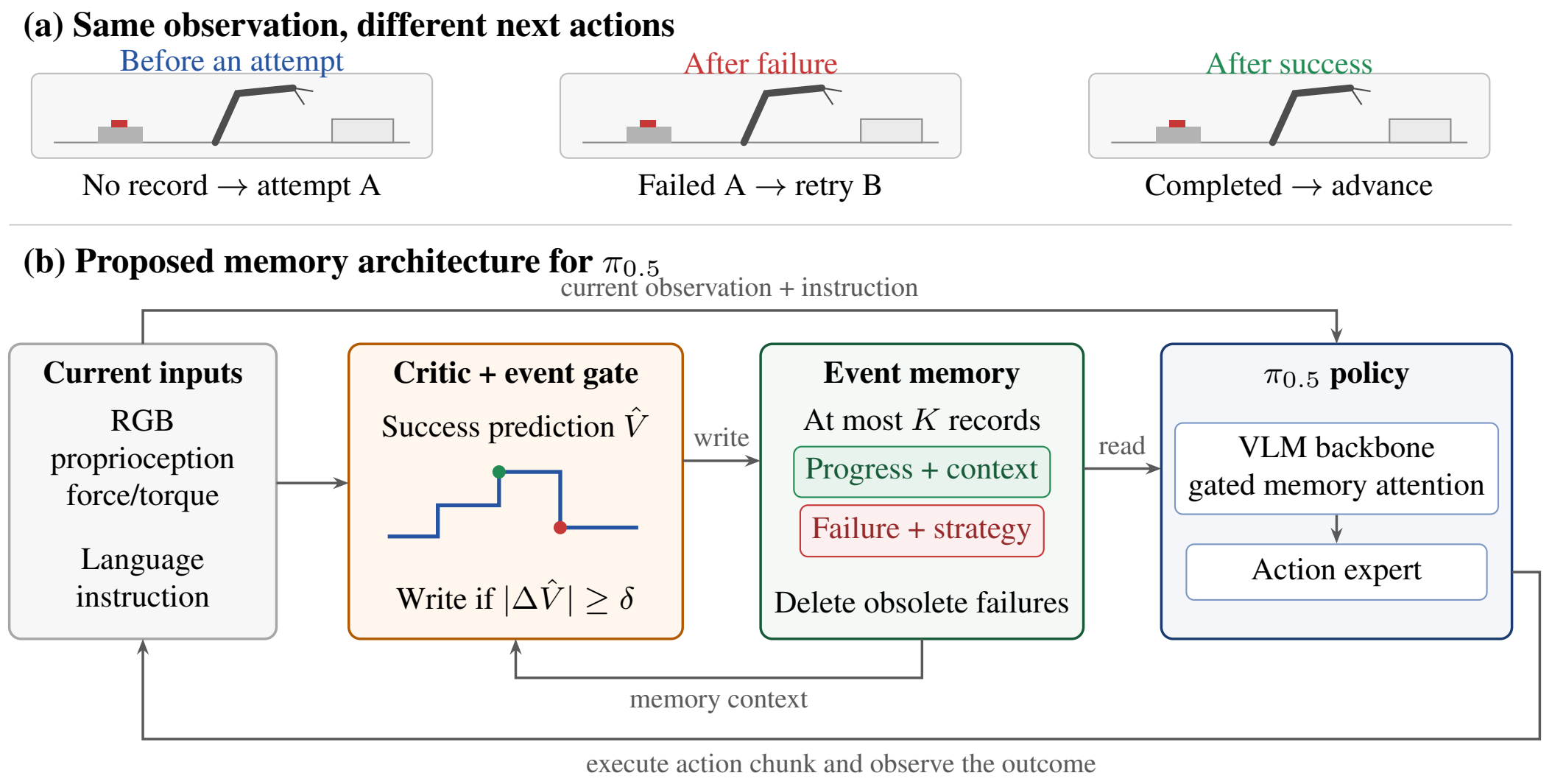

Consider a robot at its home pose beside a button (\cref{fig:system}a). The current observation may look the same before a press, after a recoverable failure, and after a successful press. Yet these situations call for different action strategies: attempt the press, retry, or advance to the next subtask. We call this ambiguity \key{decision aliasing}. Reactive policies based on robot foundation models~\cite{black2024pi0,pi2025pi05} generate actions from current observations and language instructions. When these inputs are identical, the policy cannot distinguish whether an interaction remains untried, has failed, or has succeeded.

Resolving decision aliasing therefore requires the policy to maintain a memory of its execution history. To reduce the cost of processing full observation histories, recent methods build compact memories around key events selected using perceptual, semantic, or kinematic cues~\cite{zeng2026kemo,mark2026bpp,yang2026eventvla,yang2026memorywam}. However, these cues may miss events that matter for task success. A large arm motion may be easy to detect without changing what the robot should do next. A failed contact may leave no lasting visual trace but require a different retry strategy. Selecting events for memory therefore requires more than detecting changes in appearance or motion.

To decide what to remember, we seek a signal that reflects what each event reveals about the likelihood of completing the task. Learned value estimates provide information about task progress or expected return~\cite{chen2026sarm,fang2026habc,pi2025pistar06}. We use a critic's prediction of eventual success to obtain such a signal for memory selection. A substantial change in this prediction indicates that an event has provided new information about the estimated chance of task success. This leads to our outcome-grounded principle: \emph{remember events that substantially change the predicted probability of eventual success}. 

Following this principle, we introduce \ours{} (\textbf{C}ritic-\textbf{G}ated \textbf{E}vent \textbf{M}emory), which uses a critic to select events with large estimated changes in success probability (\cref{fig:system}b). Positive and negative changes are stored as progress and failure records, respectively. These records track which subtasks have been completed and which strategies have failed. Retaining failure records allows the policy to distinguish an untried subtask from a failed attempt and adjust its retry strategy. After subtask completion, obsolete failure records are discarded.
Theoretically, under a fixed reference policy and threshold, we establish a horizon-independent bound on the expected number of large changes in conditional success probability, motivating sparse event selection. We also characterize when event records preserve the information needed for subsequent decisions.

We evaluate this mechanism in an exact-value diagnostic that isolates the effects of event selection, failure records, and consolidation. We also analyze whether learned critics on real robot data provide reliable event-detection signals and outline a proposed integration with $\pi_{0.5}$ (\cref{fig:system}b).

\textbf{Contributions.}
(i)~Theoretically, we show that the expected number of large changes in success probability has a bound independent of task length, motivating sparse event memory.
(ii)~We introduce \ours{}, a critic-gated memory of progress and failure events, and propose an architecture that integrates it with $\pi_{0.5}$.
(iii)~We evaluate \ours{} and its ablations in an exact-value diagnostic and assess learned critics for event detection on real robot data.

\section{Problem Formulation}
\label{sec:problem}

\textbf{Task and memory.}
We consider a long-horizon manipulation task with observations $o_t$, actions $a_t$, a language instruction $l$, and a terminal outcome $Z\in\{0,1\}$ that indicates task success. The execution history up to step $t$ is $h_t=(o_0,a_0,\dots,o_t)$. A policy $\pi_\theta(a_t\mid o_t,m_t,l)$ selects actions from the current observation and an event memory $m_t$ that holds at most $K$ records. Decision aliasing arises when two histories produce the same input $(o_t,m_t,l)$ but require different actions.

\textbf{Critic change as an event score.}
Whether a task succeeds depends on how the robot will act in the future, so we define success with respect to a fixed reference policy $\pi$. Let $V_t=\P_\pi(Z=1\mid h_t)$ denote the probability of eventual success given the history, and let a critic $\hat V_t$ estimate it from data. We score each transition by how much it changes this estimate:
\[
\Delta\hat V_t=\hat V_{t+1}-\hat V_t,
\qquad
\hat\kappa_t=|\Delta\hat V_t|.
\]
A large score marks a substantial change in the estimated probability of task success.

\textbf{Problem.}
Given training trajectories from $\pi$ and a memory capacity $K$, we seek a causal memory rule that allows the policy to reproduce a history-aware expert's decisions using only $(o_t,m_t,l)$. This requires histories with different expert actions to remain distinguishable through the policy input. The memory rule must determine (i)~\emph{when to write}, by selecting events using critic changes, (ii)~\emph{what to record}, including each selected event's outcome and the subtask or strategy it concerns, and (iii)~\emph{when to delete}, by identifying records that are no longer needed for subsequent decisions. All updates use only information available at the current step and maintain $|m_t|\le K$. We characterize when this goal is achievable in \cref{sec:theory} and specify three components of the memory rule in \cref{sec:method}.

\section{Theory: Sparse and Sufficient Event Memory}
\label{sec:theory}

We now justify the three decisions of \cref{sec:problem}. We analyze why changes in success probability support sparse event selection and when the resulting records preserve expert decisions. Detailed assumptions and proofs are provided in \cref{app:proofs}.

\textbf{Sparse event selection.}
Under a fixed reference policy $\pi$, the exact conditional success probability $V_t$ forms a bounded martingale~\cite{williams1991probability}. Its expected sum of squared changes is at most $1/4$, which bounds the number of changes above a fixed threshold.

\begin{proposition}[Horizon-independent event bound]
\label{prop:energy}
Let $\dV_t=V_{t+1}-V_t$. For any fixed threshold $\delta>0$ and any horizon $T$,
\[
\E_\pi\bigl|\{t<T:|\dV_t|\ge\delta\}\bigr|
\le \frac{1}{4\delta^2}.
\]
\end{proposition}

For a fixed threshold, the expected number of selected events is bounded independently of the horizon. This supports sparse event selection in long-horizon tasks.

\textbf{Records that preserve decisions.}
Sparse selection alone does not ensure that the retained events preserve all information needed for decisions. Let a \key{task register} summarize the relevant history, including completed subtasks and failed strategies, so that the register, current observation, and instruction determine the expert action. 

Our sufficiency result assumes that register updates coincide with substantial changes in success probability, so that the event-selection rule detects every update, and that each record encodes the corresponding update. Starting from a known initial register, the memory then allows its reconstruction if all required records fit within capacity $K$ and remain available. Once the relevant events have been confirmed, the expert action can be recovered from $(o_t,m_t,l)$ (\cref{thm:suff}). 


\section{\ours: Critic-Gated Event Memory}
\label{sec:method}
\ours{} selects events, records their outcomes and context, and removes obsolete records. \Cref{fig:system}b shows a proposed integration with $\pi_{0.5}$~\cite{pi2025pi05}.

\runin{When to write.}
A candidate event is written when its estimated change in success probability satisfies $\hat\kappa_t\ge\delta$. For a success-probability critic, this score is $|\hat V_{t+1}-\hat V_t|$, with predictions evaluated before and after an action chunk. The proposed architecture obtains candidates from critic-score peaks, force/torque changepoints detected by BOCPD~\cite{adams2007bocpd}, and gripper transitions. Force provides evidence for contact-only outcomes that may not appear in the image or gripper state.

\runin{What to record.}
Positive and negative estimated changes are encoded as progress and failure records. Each record also identifies the subtask or attempted strategy. In \asc{}, records contain the station, sign, and attempted variant. The proposed VLA representation combines SigLIP~\cite{zhai2023siglip} event-frame tokens with the sign and subtask or action metadata.

\runin{When to delete.}
Failure records remain available while they can affect a retry decision. Once the corresponding subtask is confirmed complete, its failure records are removed. If the bank still exceeds capacity $K$, the oldest record is evicted. The sufficiency result applies when all records needed to reconstruct the task register remain available.

\runin{Reading the memory.}
The proposed policy reads memory tokens through gated cross-attention, $X\leftarrow X+\sigma(g)X'$, where $X'$ is the attention output and $g$ is a learned gate. The critic reads the same memory to account for earlier events. Training details are given in \cref{app:vla}. The real-data analysis in \cref{app:real} separately evaluates chunk-level TD residuals from return critics as event signals.

\section{Experiments: An Exact-Value Diagnostic}
\label{sec:exp}
\begin{figure}[!t]
\centering
\begin{tikzpicture}
\begin{groupplot}[group style={group size=2 by 1, horizontal sep=1.35cm},
  annals, width=0.46\linewidth, height=4.0cm]
\nextgroupplot[xmin=1.6, xmax=10.4, ymin=-0.02, ymax=1.02,
  xtick={2,4,6,8,10}, ytick={0,0.5,1}, yticklabels={0,.5,1},
  xlabel={subtasks $N$}, ylabel={success rate}, title={(a) scaling with $N$},
  legend to name=leg:scaling, legend columns=4,
  legend style={/tikz/every even column/.append style={column sep=6pt}}]
\addplot[line width=0.4pt, mark=none] coordinates {(2,0.873) (4,0.919) (6,0.954) (8,0.967) (10,0.973)};
\addlegendentry{oracle register}
\addplot[line width=0.9pt, mark=*] coordinates {(2,0.872) (4,0.885) (6,0.916) (8,0.901) (10,0.880)};
\addlegendentry{\ours}
\addplot[dash pattern=on 3pt off 1.5pt, mark=square, mark options={solid}] coordinates {(2,0.871) (4,0.872) (6,0.915) (8,0.843) (10,0.477)};
\addlegendentry{w/o consolidation}
\addplot[densely dashed, mark=triangle, mark options={solid}] coordinates {(2,0.612) (4,0.373) (6,0.383) (8,0.255) (10,0.142)};
\addlegendentry{w/o force}
\addplot[dash pattern=on 4pt off 1pt on 1pt off 1pt, mark=diamond, mark options={solid}] coordinates {(2,0.877) (4,0.609) (6,0.600) (8,0.001) (10,0.000)};
\addlegendentry{window 36}
\addplot[dotted, line width=0.7pt, mark=x, mark options={solid}] coordinates {(2,0.477) (4,0.234) (6,0.089) (8,0.044) (10,0.017)};
\addlegendentry{w/o failure entries}
\addplot[line width=0.35pt, domain=2:10, samples=41, smooth, dash pattern=on 0.6pt off 1.2pt] {0.7^x};
\addlegendentry{ceiling $\rho^N$ (Prop.~\ref{prop:failblind})}
\nextgroupplot[xmode=log, xmin=0.001, xmax=1, ymin=0, ymax=1.02,
  xtick={0.001,0.01,0.1,1}, xticklabels={$10^{-3}$,$10^{-2}$,$10^{-1}$,$1$},
  ytick={0,0.5,1}, yticklabels={0,.5,1},
  xlabel={fraction of steps}, ylabel={share of $\sum\dV_t^2$}, title={(b) energy concentrates},
  legend style={at={(0.97,0.08)},anchor=south east,fill=white}, legend image post style={xscale=0.55}]
\addplot[dotted, line width=0.7pt, mark=o, mark size=1.1pt, mark options={solid}] coordinates {(0.001,0.261) (0.002,0.426) (0.005,0.623) (0.01,0.786) (0.02,0.874) (0.05,0.981) (0.1,0.997) (0.2,1.0) (0.5,1.0) (1,1.0)};
\addlegendentry{$N=2$}
\addplot[dashed, mark=square, mark size=1.0pt, mark options={solid}] coordinates {(0.001,0.674) (0.002,0.824) (0.005,0.894) (0.01,0.932) (0.02,0.958) (0.05,0.981) (0.1,0.995) (0.2,1.0) (0.5,1.0) (1,1.0)};
\addlegendentry{$N=6$}
\addplot[line width=0.8pt, mark=*, mark size=1.0pt] coordinates {(0.001,0.824) (0.002,0.936) (0.005,0.956) (0.01,0.965) (0.02,0.973) (0.05,0.985) (0.1,0.998) (0.2,1.0) (0.5,1.0) (1,1.0)};
\addlegendentry{$N=10$}
\end{groupplot}
\end{tikzpicture}\\[-1pt]
\pgfplotslegendfromname{leg:scaling}

\caption{\textbf{Memory scaling and sparse value changes on \asc{}.} (a)~Task success versus $N$, with the failure-blind ceiling $\rho^N$ from \cref{prop:failblind}. (b)~The top $1\%$ of steps carry $79\%$ to $96\%$ of the squared value changes, measured on 3000 reference-policy episodes per $N$.}
\label{fig:results}

\end{figure}
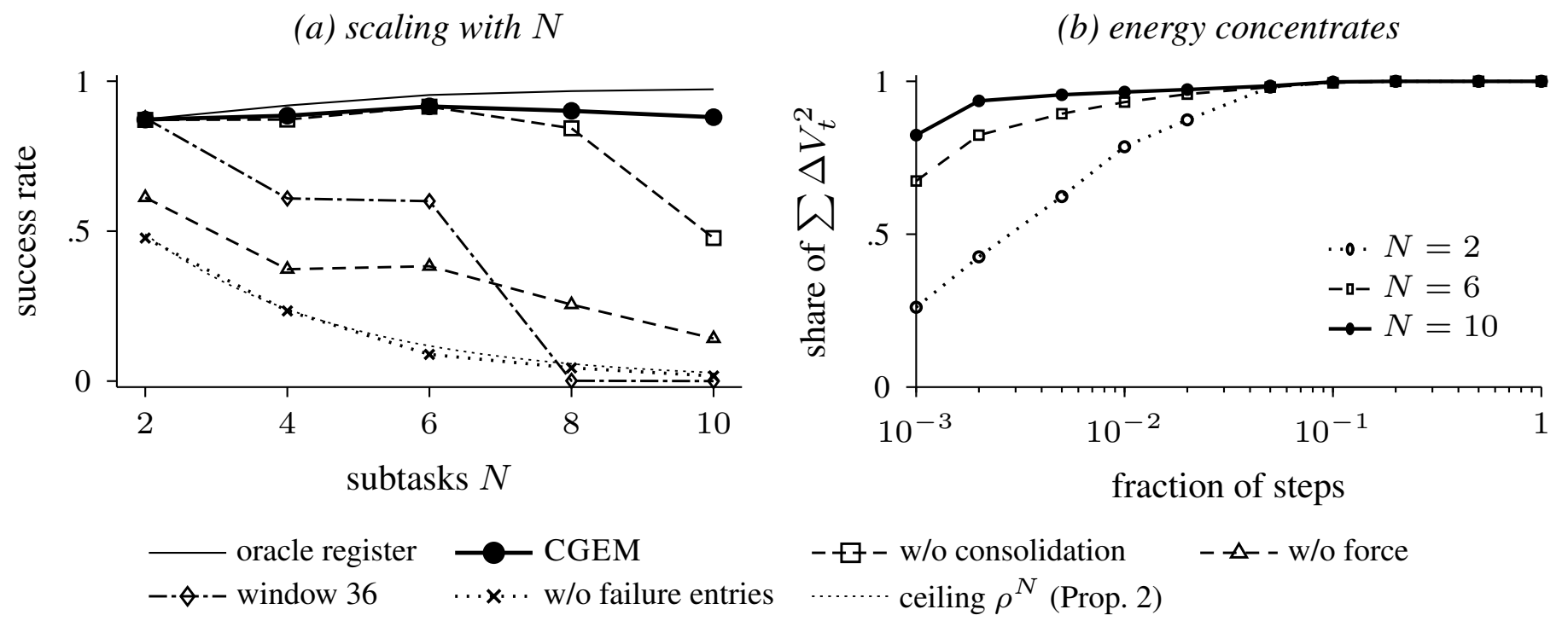

We evaluate \ours{} on the Aliased Stage Chain (\asc{}), where the robot completes $N$ ordered subtasks with repeated station visits and returns to an identical home-pose observation after each attempt. Exact success values supervise an event-score lookup fitted from local sensor features. All policies use tabular behavior cloning with DAgger~\cite{ross2011dagger}. We report success rate (SR) and stage completion (SC) over 500 episodes per seed and three seeds, with capacity $K=N$. Task and training details are in \cref{app:asc}.

\runin{Baselines.}
We compare observation windows (12 or 36 steps), uniform keyframes, kinematic events selected by gripper and visual cues~\cite{zeng2026kemo}, and an unsigned union of gripper and force events. Ablations remove failure records, force information, or consolidation. An oracle stores the task register.

\begin{table}[!htbp]
\centering\small
\setlength{\tabcolsep}{8pt}
\caption{\textbf{Task performance on \asc{}.} SR and SC are percentages. SR is reported as mean $\pm$ sample standard deviation over three seeds. Detector diagnostics are in \cref{tab:detector}.}
\label{tab:main}
\begin{tabular}{@{}lrrrr@{}}
\toprule
& \multicolumn{2}{c}{$N=6$} & \multicolumn{2}{c}{$N=10$}\\
\cmidrule(lr){2-3}\cmidrule(l){4-5}
Memory & SR & SC & SR & SC\\
\midrule
None & 0.0 & 11.6 & 0.0 & 5.6\\
Window 12 & 0.0 & 14.5 & 0.0 & 13.8\\
Window 36 & $60.0\pm3.6$ & 80.6 & 0.0 & 44.6\\
Uniform & 0.0 & 16.9 & 0.0 & 10.2\\
Kinematic & 0.0 & 11.4 & 0.0 & 5.6\\
Heuristic union & $10.0\pm0.8$ & 27.7 & $0.9\pm0.5$ & 17.3\\
\midrule
Without failure records & $8.9\pm5.6$ & 29.6 & $1.7\pm1.5$ & 16.6\\
Without force & $38.3\pm2.1$ & 46.9 & $14.2\pm0.6$ & 32.6\\
Without consolidation & $91.5\pm0.6$ & 96.2 & $47.7\pm1.1$ & 80.2\\
\textbf{\ours} & $\mathbf{91.6}\pm1.7$ & \textbf{96.3} & $\mathbf{88.0}\pm0.9$ & \textbf{93.6}\\
Oracle register & $95.4\pm0.4$ & 99.0 & $97.3\pm0.3$ & 99.4\\
\bottomrule
\end{tabular}
\end{table}

\runin{Main findings.}
\ours{} maintains high success as task length increases (\cref{fig:results}a). At $N=10$, it achieves $88.0\%$ SR, while the observation windows, kinematic memory, and uniform memory all reach $0.0\%$ (\cref{tab:main}).

The ablations show that event selection, outcome records, and consolidation play complementary roles. Removing failure records reduces SR to $1.7\%$ at $N=10$. Without force information, sign accuracy falls despite a similar share of retained squared value changes (\cref{tab:detector}), highlighting the need to identify event outcomes. Removing obsolete failure records raises SR from $47.7\%$ to $88.0\%$ at $N=10$, supporting consolidation under limited capacity.

Squared value changes concentrate in a small fraction of steps (\cref{fig:results}b), consistent with sparse event selection. \Cref{app:more} provides additional diagnostics, and \cref{app:real} evaluates event signals from learned critics on real robot data.

\FloatBarrier
\bibliographystyle{plainnat}
\bibliography{cgem}

\clearpage
\appendix

\section{Formal Statements and Proofs}
\label{app:proofs}

\subsection{Energy identity and event-count bound}
Fix the reference policy $\pi$ and instruction $l$, and let $\F_t=\sigma(h_t,l)$. Since $V_t=\E[Z\mid\F_t]$ and $Z\in\{0,1\}$, $(V_t)$ is a bounded martingale. Its increments are orthogonal in expectation. For a finite horizon $T$,
\[
\E\sum_{t<T}\dV_t^2=\E[V_T^2]-\E[V_0^2]
\le\E[V_0(1-V_0)]\le\frac14.
\]
The same bound holds for a bounded stopping time by applying this identity to the stopped process. When the terminal outcome has been observed, $V_T=Z$, the first inequality is an equality. With a fixed initial value $V_0=p$, this gives $\E\sum_{t<T}\dV_t^2=p(1-p)$.

\begin{proof}[Proof of \cref{prop:energy}]
Since $\delta^2\1\{|\dV_t|\ge\delta\}\le\dV_t^2$, summing over the episode and taking expectations gives
\[
\E\bigl|\{t<T:|\dV_t|\ge\delta\}\bigr|
\le\frac{1}{\delta^2}\E\sum_{t<T}\dV_t^2\le\frac{1}{4\delta^2}.
\]
\end{proof}

\subsection{Records that preserve expert decisions}
\begin{theorem}[Conditional sufficiency of event records]
\label{thm:suff}
Suppose a task register $s_t$ has a known initial value $s_0(l)$ and determines the expert action through $a_t^\star=\mu(s_t,o_t,l)$. Every register change is detected and encoded in a record $e$ with a fixed update map $s_{t+1}=U(s_t,e)$. Records are available after confirmation and retain their event order. If all records needed for reconstruction fit within capacity $K$ and are retained, then, at times with no pending updates, a fixed decoder recovers $s_t=D(m_t,l)$. The expert action is therefore a function of $(o_t,m_t,l)$. A record can be deleted if doing so leaves the decoded register unchanged.
\end{theorem}
\begin{proof}
Starting from $s_0(l)$, apply $U$ to the confirmed records in event order. Each record reproduces its register update, and the register is unchanged between events. Induction reconstructs the current register once all relevant updates have been confirmed. Retaining the required records and deleting only records that leave the decoded register unchanged preserve this reconstruction. Substituting $D(m_t,l)$ into $\mu$ recovers the expert action.
\end{proof}
For critic-based selection, detecting every required update is an additional condition. One sufficient case is that all register updates occur at candidate steps with $|\dV_t|\ge\delta$, while other candidates remain below the threshold. Records must encode the outcome and strategy information needed for the update. If subtask completion resets the register's failure history, its completion record permits those failure records to be deleted.

\subsection{Failure-blind success ceiling}
\begin{proposition}[Failure-blind ceiling]
\label{prop:failblind}
Consider $N$ sequential subtasks with independent latent correct strategies $b_i\in\{A,B\}$ and $\max_b\P(b_i=b)=\rho$. Before its first attempt, the policy has no information about $b_i$ beyond this prior. Only the correct strategy can complete the subtask. If a failed attempt leaves the entire input at the next strategy decision unchanged, a deterministic policy acting on that input succeeds with probability at most $\rho^N$.
\end{proposition}
\begin{proof}
The unchanged decision input makes the policy repeat its initial strategy $\hat b_i$ on every retry. Completing subtask $i$ therefore requires $\hat b_i=b_i$, which has conditional probability at most $\rho$, by independence and the prior assumption. Multiplying across subtasks gives $\P(\mathrm{success})\le\rho^N$. Slips or a finite time budget can further reduce success.
\end{proof}

\section{The Aliased Stage Chain}
\label{app:asc}
\textbf{Task.} There are three stations $\{1,2,3\}$ of types P, G, and G. The plan is the length-$N$ prefix of $(1,2,1,3,2,1,3,1,2,3)$, so stations repeat and progress must be counted. Each attempt has four phases. (i)~At \emph{home}, the robot chooses a station, and choosing a wrong station is an irrecoverable ordering error. (ii)~\emph{Outbound}: the robot makes $L=6$ moves, each of which stalls with probability $0.2$, and chooses a variant on arrival. (iii)~\emph{Interaction}: after $D=3$ steps, the attempt succeeds with probability $1-q$, where $q=0.1$, if the variant equals the latent $b_i$, where $\P(b_i\!=\!A)=\rho=0.7$, and fails otherwise. (iv)~\emph{Return}: the robot makes $L$ moves back to home. The time limit is $40N$ steps.

\textbf{Observations.} Phase, position, and current station are always observed, and the home observation is identical across visits. Force is $\mathcal N(0,0.1^2)$ while moving and $\mathcal N(0.6,0.15^2)$ or $\mathcal N(0.3,0.1^2)$ during P or G interaction. At the outcome step, a P success gives $\mathcal N(2.5,0.3^2)$ and a P failure $\mathcal N(1.2,0.3^2)$. A G success gives $\mathcal N(0.9,0.2^2)$ with the gripper opening and the visual cue ``placed'', and a G failure gives $\mathcal N(0.05,0.1^2)$ with the gripper opening and the cue ``dropped''. Force is discretized into five bins.

\textbf{Exact $V_t$.} The reference policy is the history-aware Bayes expert. It chooses the correct station with probability $1-\varepsilon_h$ ($\varepsilon_h=0.02$) and the posterior-MAP variant with probability $1-\varepsilon_v$ ($\varepsilon_v=0.05$), where $\P(b_i\!=\!A\mid n_A,n_B)=\rho q^{n_A}/\big(\rho q^{n_A}+(1-\rho)q^{n_B}\big)$ and $n_A$ and $n_B$ count failed attempts with each variant. The finite diagnostic value computation caps failure counts at 3 and computes $V_t$ by backward induction on a sparse transition matrix.

\textbf{Learning.} The keyness head is a lookup table of $\E|\dV_t|$ and $\E\dV_t$ given local features (station, phase, gripper event, visual cue, and force bin), fitted on 1500 reference rollouts. A record is written when $\hat\kappa_t\ge\delta=0.01$, with a refractory period of 2 steps. Records are (station, sign, own variant). When a progress record is written, all failure records are deleted, and otherwise the bank is FIFO with capacity $K$. Policies use tabular behavior cloning with four DAgger rounds (mixing coefficients $1,.5,.25,0$ and 300 episodes per round). Their keys are the ordered records at home and the station together with the last three records on arrival, and unseen keys back off to the per-station majority action.

\section{Additional Results}
\label{app:more}
\runin{Detector diagnostics.}
\Cref{tab:detector} reports the energy retained by each detector and the accuracy of its event signs. \Cref{fig:teaser} illustrates the signed records along a reference-policy episode.

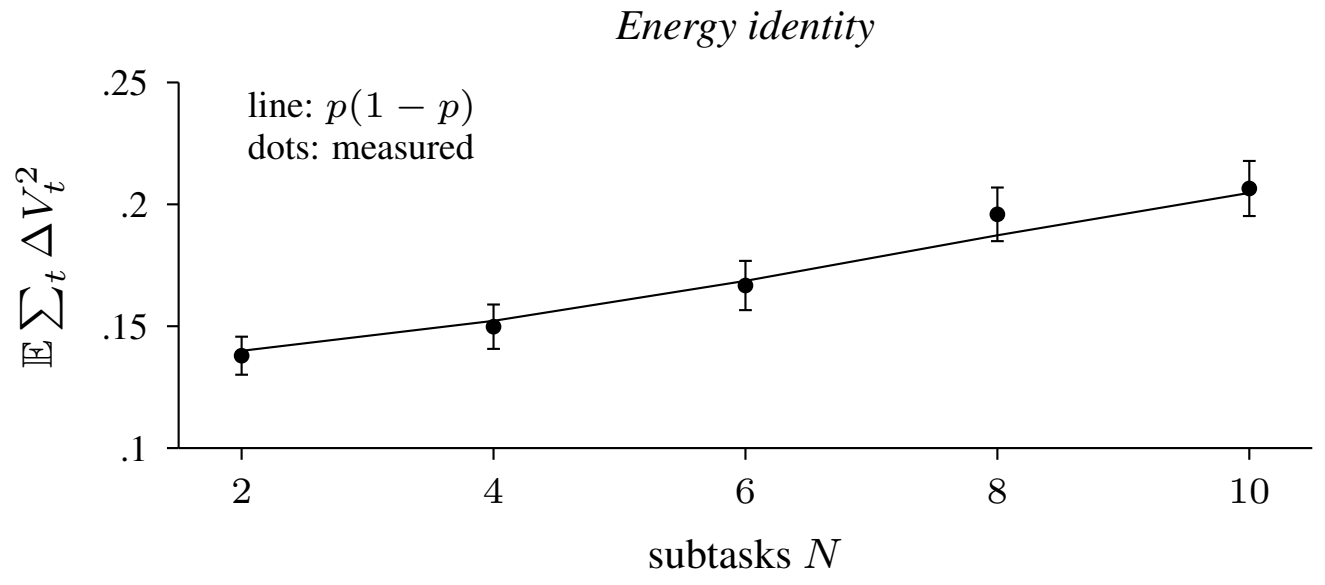
\begin{figure}[!htbp]
\centering
\begin{tikzpicture}
\begin{axis}[annals,width=.65\linewidth,height=4cm,xmin=1.5, xmax=10.5, ymin=0.1, ymax=0.25,
  xtick={2,4,6,8,10}, ytick={0.1,0.15,0.2,0.25}, yticklabels={.1,.15,.2,.25},
  xlabel={subtasks $N$}, ylabel={$\E\sum_t\dV_t^2$}, title={Energy identity}]
\addplot[line width=0.4pt, mark=none] coordinates {(2,0.1399) (4,0.1522) (6,0.1686) (8,0.1873) (10,0.2047)};
\addplot[only marks, mark=*, mark size=1.2pt, error bars/.cd, y dir=both, y explicit] coordinates {(2,0.1379) +- (0,0.0078) (4,0.1498) +- (0,0.0091) (6,0.1667) +- (0,0.0101) (8,0.1959) +- (0,0.0110) (10,0.2065) +- (0,0.0113)};
\node[font=\scriptsize, anchor=north west, align=left, inner sep=1pt] at (axis cs:2,0.25) {line: $p(1-p)$\\dots: measured};

\end{axis}

\end{tikzpicture}
\caption{\textbf{Energy-identity check.} The line shows $p(1-p)$ and the points show estimates of $\E\sum_t\dV_t^2$ from 3000 reference-policy episodes per $N$, with error bars of $\pm2$ standard errors.}
\label{fig:energy}
\end{figure}

\begin{table}[!htbp]
\centering\small
\caption{\textbf{Detector diagnostics at $N=6$.} E-rec is the percentage of $\sum_t\dV_t^2$ on written steps. Sign is the accuracy (\%) of stored progress/failure labels.}
\label{tab:detector}
\begin{tabular}{@{}lrr@{}}
\toprule Memory & E-rec & Sign\\\midrule
Uniform & 3.6 & --\\
Kinematic & 12.1 & 100\\
Heuristic union & 60.3 & n/a\\
Without failure records & 28.4 & 99.4\\
Without force & 85.8 & 68.3\\
Without consolidation & 85.8 & 99.3\\
\ours{} & 85.8 & 99.3\\\bottomrule
\end{tabular}
\end{table}
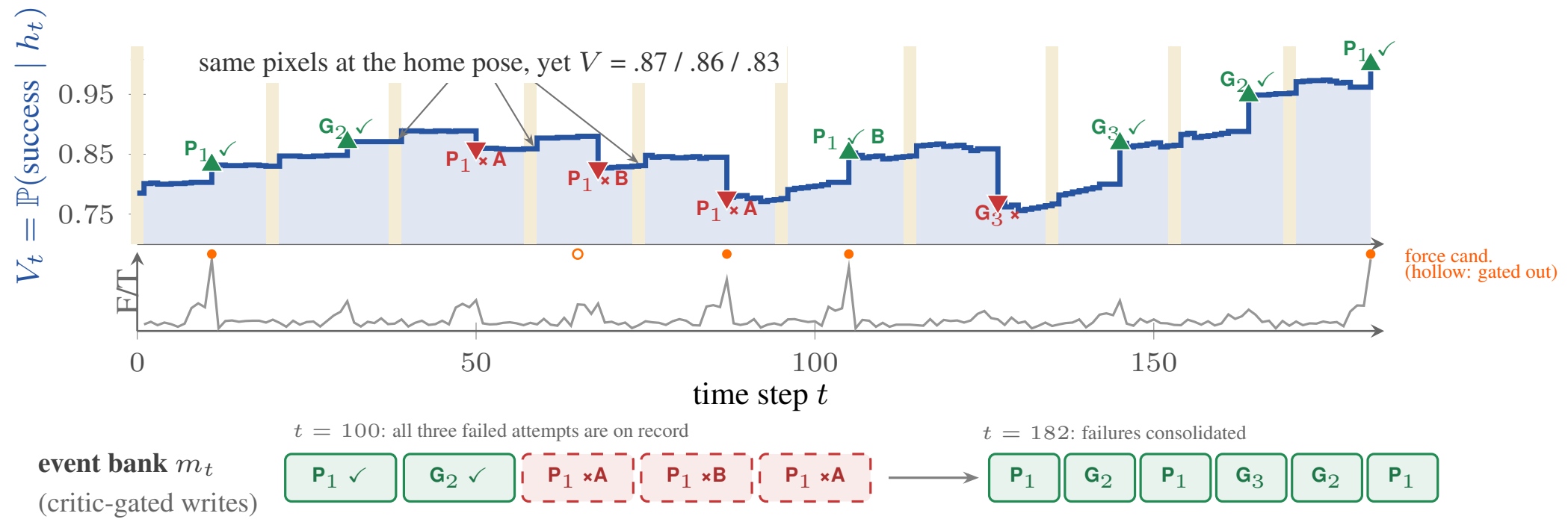
\begin{figure}[!htbp]
\centering
\begin{tikzpicture}
\begin{axis}[
  name=vax,
  width=0.79\linewidth, height=1.75cm,
  scale only axis,
  xmin=0, xmax=184, ymin=0.70, ymax=1.03,
  axis lines=left,
  axis line style={draw=black!60, line width=0.5pt},
  xtick=\empty,
  ytick={0.75,0.85,0.95},
  tick label style={font=\scriptsize, text=black!70},
  ylabel={\footnotesize $V_t=\P(\text{success}\mid h_t)$},
  ylabel style={yshift=-4pt, text=cV},
  clip=false,
]
\foreach \x in {0,20,38,58,74,95,114,135,153,170}{
  \edef\tmp{\noexpand\fill[cAlias] (axis cs:\x-0.9,0.70) rectangle (axis cs:\x+0.9,1.03);}\tmp
}
\addplot[name path=V, cV, line width=1.3pt, const plot] coordinates {(0,0.785) (1,0.801) (2,0.802) (3,0.800) (4,0.800) (5,0.801) (6,0.801) (7,0.802) (8,0.803) (9,0.803) (10,0.803) (11,0.832) (12,0.832) (13,0.831) (14,0.831) (15,0.831) (16,0.832) (17,0.832) (18,0.831) (19,0.830) (20,0.830) (21,0.847) (22,0.847) (23,0.847) (24,0.846) (25,0.846) (26,0.847) (27,0.847) (28,0.848) (29,0.848) (30,0.848) (31,0.870) (32,0.871) (33,0.871) (34,0.871) (35,0.871) (36,0.871) (37,0.871) (38,0.871) (39,0.889) (40,0.889) (41,0.889) (42,0.888) (43,0.888) (44,0.889) (45,0.889) (46,0.888) (47,0.889) (48,0.889) (49,0.889) (50,0.860) (51,0.860) (52,0.860) (53,0.859) (54,0.858) (55,0.858) (56,0.858) (57,0.859) (58,0.859) (59,0.877) (60,0.877) (61,0.877) (62,0.878) (63,0.878) (64,0.878) (65,0.880) (66,0.880) (67,0.880) (68,0.827) (69,0.827) (70,0.828) (71,0.829) (72,0.829) (73,0.830) (74,0.831) (75,0.848) (76,0.845) (77,0.846) (78,0.846) (79,0.844) (80,0.844) (81,0.845) (82,0.846) (83,0.843) (84,0.845) (85,0.845) (86,0.845) (87,0.778) (88,0.780) (89,0.781) (90,0.776) (91,0.777) (92,0.771) (93,0.773) (94,0.774) (95,0.776) (96,0.791) (97,0.793) (98,0.794) (99,0.796) (100,0.797) (101,0.799) (102,0.803) (103,0.803) (104,0.803) (105,0.852) (106,0.848) (107,0.844) (108,0.846) (109,0.847) (110,0.842) (111,0.843) (112,0.845) (113,0.846) (114,0.847) (115,0.864) (116,0.865) (117,0.866) (118,0.867) (119,0.863) (120,0.864) (121,0.865) (122,0.861) (123,0.856) (124,0.859) (125,0.859) (126,0.859) (127,0.771) (128,0.762) (129,0.765) (130,0.756) (131,0.758) (132,0.760) (133,0.762) (134,0.764) (135,0.767) (136,0.782) (137,0.784) (138,0.787) (139,0.789) (140,0.792) (141,0.794) (142,0.800) (143,0.800) (144,0.800) (145,0.868) (146,0.862) (147,0.864) (148,0.865) (149,0.867) (150,0.869) (151,0.862) (152,0.864) (153,0.865) (154,0.883) (155,0.885) (156,0.878) (157,0.880) (158,0.881) (159,0.883) (160,0.885) (161,0.888) (162,0.888) (163,0.888) (164,0.948) (165,0.949) (166,0.949) (167,0.950) (168,0.951) (169,0.951) (170,0.952) (171,0.971) (172,0.972) (173,0.973) (174,0.973) (175,0.974) (176,0.972) (177,0.969) (178,0.970) (179,0.962) (180,0.962) (181,0.962) (182,1.000)};
\path[name path=B] (axis cs:0,0.70) -- (axis cs:182,0.70);
\addplot[cV!14] fill between[of=V and B];
\addplot[only marks, mark=triangle*, mark size=3.2pt, cPos, mark options={draw=white, line width=0.3pt}]
  coordinates {(11,0.832) (31,0.870) (105,0.852) (145,0.868) (164,0.948) (182,1.000)};
\addplot[only marks, mark=triangle*, mark size=3.2pt, cNeg, mark options={rotate=180, draw=white, line width=0.3pt}]
  coordinates {(50,0.860) (68,0.827) (87,0.778) (127,0.771)};
\node[font=\tiny\sffamily\bfseries, text=cPos, inner sep=1pt] at (axis cs:11,0.855) {P$_1$\,\checkmark};
\node[font=\tiny\sffamily\bfseries, text=cPos, inner sep=1pt] at (axis cs:31,0.893) {G$_2$\,\checkmark};
\node[font=\tiny\sffamily\bfseries, text=cNeg, inner sep=1pt] at (axis cs:50,0.838) {P$_1$\,\texttimes\,A};
\node[font=\tiny\sffamily\bfseries, text=cNeg, inner sep=1pt] at (axis cs:68,0.805) {P$_1$\,\texttimes\,B};
\node[font=\tiny\sffamily\bfseries, text=cNeg, inner sep=1pt] at (axis cs:87,0.755) {P$_1$\,\texttimes\,A};
\node[font=\tiny\sffamily\bfseries, text=cPos, inner sep=1pt] at (axis cs:105,0.875) {P$_1$\,\checkmark\,B};
\node[font=\tiny\sffamily\bfseries, text=cNeg, inner sep=1pt] at (axis cs:127,0.748) {G$_3$\,\texttimes};
\node[font=\tiny\sffamily\bfseries, text=cPos, inner sep=1pt] at (axis cs:145,0.891) {G$_3$\,\checkmark};
\node[font=\tiny\sffamily\bfseries, text=cPos, inner sep=1pt] at (axis cs:164,0.971) {G$_2$\,\checkmark};
\node[font=\tiny\sffamily\bfseries, text=cPos, inner sep=1pt] at (axis cs:182,1.022) {P$_1$\,\checkmark};
\draw[black!55, line width=0.4pt, -{Stealth[length=3pt]}] (axis cs:47,0.985) -- (axis cs:38.5,0.875);
\draw[black!55, line width=0.4pt, -{Stealth[length=3pt]}] (axis cs:52,0.985) -- (axis cs:58.5,0.863);
\draw[black!55, line width=0.4pt, -{Stealth[length=3pt]}] (axis cs:57,0.985) -- (axis cs:74,0.835);
\node[font=\scriptsize, align=center, fill=white, inner sep=1.5pt, text=black!80] at (axis cs:52,1.0)
  {same pixels at the home pose, yet $V$ = .87 / .86 / .83};
\end{axis}
\begin{axis}[
  at={(vax.south west)}, anchor=north west, yshift=-2pt,
  width=0.79\linewidth, height=0.7cm, scale only axis,
  xmin=0, xmax=184, ymin=-0.3, ymax=2.7,
  axis lines=left, axis line style={draw=black!60, line width=0.5pt},
  ytick=\empty, xtick={0,50,100,150},
  tick label style={font=\scriptsize, text=black!70},
  xlabel={\footnotesize time step $t$}, xlabel style={yshift=4pt},
  ylabel={\footnotesize F/T}, ylabel style={yshift=-6pt, text=cF!70!black},
  clip=false,
]
\addplot[cF, line width=0.6pt] coordinates {(1,-0.05) (2,0.05) (3,-0.08) (4,0.02) (5,-0.04) (6,-0.14) (7,-0.02) (8,0.57) (9,0.75) (10,0.60) (11,2.40) (12,-0.20) (13,0.08) (14,0.10) (15,0.05) (16,0.08) (17,0.09) (18,0.09) (19,0.02) (20,0.13) (21,0.01) (22,-0.14) (23,0.09) (24,-0.06) (25,0.04) (26,0.06) (27,-0.13) (28,0.08) (29,0.29) (30,0.43) (31,0.83) (32,0.02) (33,-0.13) (34,0.05) (35,-0.00) (36,0.18) (37,-0.04) (38,-0.02) (39,0.14) (40,-0.13) (41,0.01) (42,0.09) (43,-0.07) (44,-0.04) (45,0.09) (46,-0.05) (47,0.59) (48,0.73) (49,0.60) (50,0.87) (51,-0.01) (52,-0.05) (53,-0.04) (54,0.01) (55,0.03) (56,-0.01) (57,-0.08) (58,0.01) (59,-0.05) (60,0.16) (61,-0.06) (62,-0.10) (63,0.20) (64,-0.10) (65,0.71) (66,0.69) (67,0.35) (68,0.74) (69,0.02) (70,0.05) (71,-0.05) (72,0.02) (73,-0.19) (74,-0.08) (75,0.07) (76,0.03) (77,0.01) (78,-0.07) (79,0.07) (80,0.15) (81,0.09) (82,-0.08) (83,-0.09) (84,0.63) (85,0.68) (86,0.61) (87,1.65) (88,-0.01) (89,-0.07) (90,-0.02) (91,0.15) (92,-0.17) (93,-0.06) (94,-0.01) (95,-0.08) (96,0.14) (97,0.15) (98,0.09) (99,0.04) (100,0.07) (101,-0.10) (102,0.62) (103,0.47) (104,0.56) (105,2.10) (106,-0.20) (107,0.12) (108,0.09) (109,-0.14) (110,0.01) (111,-0.14) (112,-0.06) (113,-0.06) (114,-0.06) (115,-0.08) (116,-0.10) (117,0.15) (118,-0.09) (119,0.12) (120,0.08) (121,0.10) (122,0.02) (123,0.10) (124,0.35) (125,0.46) (126,0.19) (127,0.14) (128,-0.07) (129,0.21) (130,0.00) (131,0.06) (132,-0.20) (133,-0.10) (134,0.02) (135,-0.11) (136,-0.05) (137,-0.00) (138,0.02) (139,0.14) (140,-0.16) (141,0.02) (142,0.32) (143,0.37) (144,0.30) (145,0.85) (146,-0.02) (147,-0.10) (148,0.16) (149,0.02) (150,0.14) (151,-0.00) (152,-0.01) (153,-0.07) (154,-0.08) (155,0.13) (156,0.05) (157,-0.06) (158,0.07) (159,-0.09) (160,-0.01) (161,0.33) (162,0.16) (163,0.26) (164,0.51) (165,0.32) (166,0.04) (167,0.09) (168,-0.20) (169,0.08) (170,-0.10) (171,0.06) (172,-0.08) (173,-0.01) (174,-0.02) (175,0.06) (176,0.01) (177,0.04) (178,-0.12) (179,0.44) (180,0.67) (181,0.66) (182,2.42)};
\foreach \x in {11,87,105,182}{\edef\tmp{\noexpand\fill[orange!85!red] (axis cs:\x,2.62) circle (1.2pt);}\tmp}
\draw[orange!85!red, line width=0.5pt] (axis cs:65,2.62) circle (1.2pt);
\node[font=\tiny, text=orange!70!red, anchor=west, align=left] at (axis cs:185,2.2) {force cand.\\[-2pt](hollow: gated out)};
\end{axis}
\begin{scope}[shift={($(vax.south west)+(-1.0cm,-1.95cm)$)}]
  \node[anchor=west, font=\scriptsize\bfseries, text=black!75] at (0,0) {event bank $m_t$};
  \node[anchor=west, font=\scriptsize, text=black!60] at (0,-0.36) {(critic-gated writes)};
  \tikzset{slot/.style={draw=#1, fill=#1!10, rounded corners=1.6pt, minimum height=0.42cm, minimum width=0.98cm, font=\tiny\sffamily\bfseries, text=#1!85!black, line width=0.6pt}}
  \node[font=\tiny, text=black!60, anchor=west] at (2.25,0.28) {$t=100$: all three failed attempts are on record};
  \node[slot=cPos] at (2.80,-0.12) {P$_1$ \checkmark};
  \node[slot=cPos] at (3.85,-0.12) {G$_2$ \checkmark};
  \node[slot=cNeg, dashed] at (4.90,-0.12) {P$_1$ \texttimes A};
  \node[slot=cNeg, dashed] at (5.95,-0.12) {P$_1$ \texttimes B};
  \node[slot=cNeg, dashed] at (7.00,-0.12) {P$_1$ \texttimes A};
  \draw[-{Stealth[length=4pt]}, black!50, line width=0.6pt] (7.65,-0.12) -- (8.45,-0.12);
  \node[font=\tiny, text=black!60, anchor=west] at (8.35,0.28) {$t=182$: failures consolidated};
  \node[slot=cPos, minimum width=0.62cm] at (8.85,-0.12) {P$_1$};
  \node[slot=cPos, minimum width=0.62cm] at (9.52,-0.12) {G$_2$};
  \node[slot=cPos, minimum width=0.62cm] at (10.19,-0.12) {P$_1$};
  \node[slot=cPos, minimum width=0.62cm] at (10.86,-0.12) {G$_3$};
  \node[slot=cPos, minimum width=0.62cm] at (11.53,-0.12) {G$_2$};
  \node[slot=cPos, minimum width=0.62cm] at (12.20,-0.12) {P$_1$};
\end{scope}
\end{tikzpicture}
\vspace{-2pt}
\caption{\textbf{What should a robot remember? The jumps in its odds.} P denotes a contact-only press and G a grasp-and-place, and subscripts index stations. The exact $V_t$ jumps at \textcolor{cPos}{\textbf{progress}} and \textcolor{cNeg}{\textbf{failure}} outcomes, while the home-pose observation (shaded) never changes. \ours{} writes the signed outcome jumps and skips the small changes when the robot leaves home. It keeps failure records until the subtask succeeds, which allows the switch to variant B at $t\!=\!105$, and then deletes them.}
\label{fig:teaser}

\end{figure}
\runin{Energy identity.}
The empirical energy estimates agree with $p(1-p)$ within two standard errors over mean episode lengths from 47 to 243 steps (\cref{fig:energy,tab:more}). Changes larger than $0.05$ occur $1.2$ to $3.1$ times per episode over this range.

\begin{table}[!htbp]
\centering\small
\setlength{\tabcolsep}{5pt}
\caption{Top: energy identity (\cref{prop:energy}), where $\bar T$ is the mean horizon and $\pm$ is one standard error. Bottom: SR (\%) at $N\!=\!6$ versus capacity $K$.}
\label{tab:more}
\begin{tabular}{@{}ccccc@{}}
\toprule
$N$ & $\bar T$ & $p(1-p)$ & $\E\sum\dV_t^2$ & \#$\{\kappa_t\!\ge\!.05\}$ \\
\midrule
2 & 47.3 & .1399 & .1379{\scriptsize$\pm$.0039} & 2.49 \\
4 & 102.5 & .1522 & .1498{\scriptsize$\pm$.0045} & 3.13 \\
6 & 152.4 & .1686 & .1667{\scriptsize$\pm$.0051} & 2.32 \\
8 & 199.9 & .1873 & .1959{\scriptsize$\pm$.0055} & 1.87 \\
10 & 243.2 & .2047 & .2065{\scriptsize$\pm$.0056} & 1.23 \\
\bottomrule
\end{tabular}\par\medskip
\begin{tabular}{@{}lccccccc@{}}
\toprule
$K$ & 1 & 2 & 3 & 4 & 5 & 6 & 8 \\
\midrule
kinematic & 0.0 & 0.0 & 0.0 & 0.0 & 0.0 & 0.0 & 0.0 \\
w/o cons. & 0.0 & 58.0 & 84.5 & 89.2 & 90.4 & 91.5 & 87.5 \\
\ours & 0.0 & 82.7 & 89.6 & 91.1 & 91.3 & 91.6 & 91.9 \\
\bottomrule
\end{tabular}
\end{table}
\runin{Capacity.} At $N=6$, \ours{} reaches $82.7\%$ SR with $K=2$ and $91.6\%$ with $K=6$, compared with $58.0\%$ and $91.5\%$ without consolidation. All three tested memory variants have zero SR with $K=1$ (\cref{tab:more}).

\clearpage
\section{Proposed VLA Integration}
\label{app:vla}
The proposed architecture (\cref{fig:system}) would replay the causal memory rule over demonstrations and robot rollouts to construct inputs for flow-matching fine-tuning. Subtask annotations and episode outcomes would supervise the critic, with outcome labels restricted to autonomous segments at intervention boundaries~\cite{fang2026habc}.

An optional extension weights imitation by $w_t=1+\lambda|\hat A_t|$, where $\hat A_t$ estimates action advantage. This extension is separate from the memory mechanism evaluated in \asc{}. Planned evaluation on LIBERO-Long and real-robot tasks would compare task success and repeated unsuccessful retries against the base policy, observation windows, and event-memory baselines.

\section{Event Detection by Learned Critics on Real Bimanual Data}
\label{app:real}
We evaluate five learned critics as failure-event detectors on a real dual-arm task (AgiBot charging: pick the left charger head, insert it, pick the right one, and insert it), recorded at 30\,fps with $\gamma=0.999$ per frame and action chunks of $H=50$ frames. The critics are DINOv2-based~\cite{oquab2023dinov2} twin-$Q$/$V$ networks trained by Monte Carlo regression (MC) or IQL~\cite{kostrikov2022iql} under three rewards.
\begin{itemize}[leftmargin=*,itemsep=0pt,topsep=1pt]
\item \emph{Mode A}: outcome only, with $+1$ on the success frame.
\item \emph{Mode 7}: milestones and a regression penalty predicted by a VLM reward model, plus a time cost, with no human failure labels.
\item \emph{Mode 9}: human-annotated milestones plus $-0.25$ at the start of every \emph{human-annotated failure window}.
\end{itemize}
Evaluation uses all 87 held-out test episodes, none of which were seen during training or checkpoint selection. They comprise 75 expert episodes and 12 suboptimal episodes with 42 annotated recoverable failures (16 failed insertions, 15 empty grasps, and 11 bad grasp poses). All episodes end in success, so failures are evaluated as windows within episodes.

\textbf{Score.} These critics estimate discounted returns under their respective rewards. We evaluate the chunk-level TD residual $d_t=\sum_{k<H}\gamma^kr_{t+k}+\gamma^H\hat V_{t+H}-\hat V_t$, smoothed with a seven-point moving average. Its value-only part $\gamma^H\hat V_{t+H}-\hat V_t$ excludes the in-chunk reward.

\textbf{Metrics.} $\kappa_0$ is the 99th percentile of $|d_t|$ over all expert steps (26{,}846 steps, 74.6\,min). $\kappa_{\min}$ is the 10th percentile, over failure windows, of the peak of $-d_t$ within each window extended by $\pm1$\,s. A window counts as detected if $d_t\le-\theta$ somewhere inside it, for a threshold $\theta$. False writes are threshold crossings on expert episodes, with crossings within 0.3\,s merged.

\begin{table}[!htbp]
\centering\small
\setlength{\tabcolsep}{4pt}
\caption{Event-detection metrics of real critics (smoothed $d_t$, 87 test episodes, 42 failure windows). ``Labels in reward'' marks critics whose reward contains the evaluation annotations. AUROC is per step, with $-d_t$ scoring steps inside annotated windows against all other test steps.}
\label{tab:realgap}
\begin{tabular}{@{}lcccccccc@{}}
\toprule
 & labels & & & \multicolumn{2}{c}{margin $\kappa_{\min}-\kappa_0$} & & \multicolumn{2}{c}{$\theta=0.1$} \\
\cmidrule(lr){5-6}\cmidrule(l){8-9}
Critic & in reward & $\kappa_0$ & $\kappa_{\min}$ & $d_t$ & value only & AUROC & recall & FP/min \\
\midrule
MC Mode A @1k & no & .066 & .015 & $-.051$ & $-.051$ & .71 & .12 & .07 \\
MC Mode 7 @2k & no & .098 & .081 & $-.017$ & $-.076$ & .83 & .74 & .43 \\
IQL Mode 7 @6.5k & no & .084 & .087 & $+.003$ & $-.079$ & \textbf{.86} & .76 & .12 \\
MC Mode 9 @2k & yes & .080 & .191 & $+.111$ & $-.071$ & .74 & 1.00 & .01 \\
IQL Mode 9 @7k & yes & .066 & .221 & $+.155$ & $-.071$ & .72 & 1.00 & .04 \\
\bottomrule
\end{tabular}
\end{table}

\begin{figure}[p]
\centering

\begin{tikzpicture}[font=\scriptsize]
\foreach \x/\y/\c/\t in {0/0/red/{MC Mode 7 @2k},4.2/0/orange/{MC Mode 9 @2k},8.4/0/green!60!black/{IQL Mode 9 @7k},0/-.38/blue/{IQL Mode 7 @6.5k},4.2/-.38/black/{MC Mode A @1k}}{
\draw[\c,line width=.9pt] (\x,\y)--(\x+.35,\y);
\node[anchor=west] at (\x+.42,\y) {\t};
}
\end{tikzpicture}

\begin{tikzpicture}
\node[inner sep=0] (rocplot) {\includegraphics[width=.73\linewidth,trim=365bp 0bp 0bp 0bp,clip]{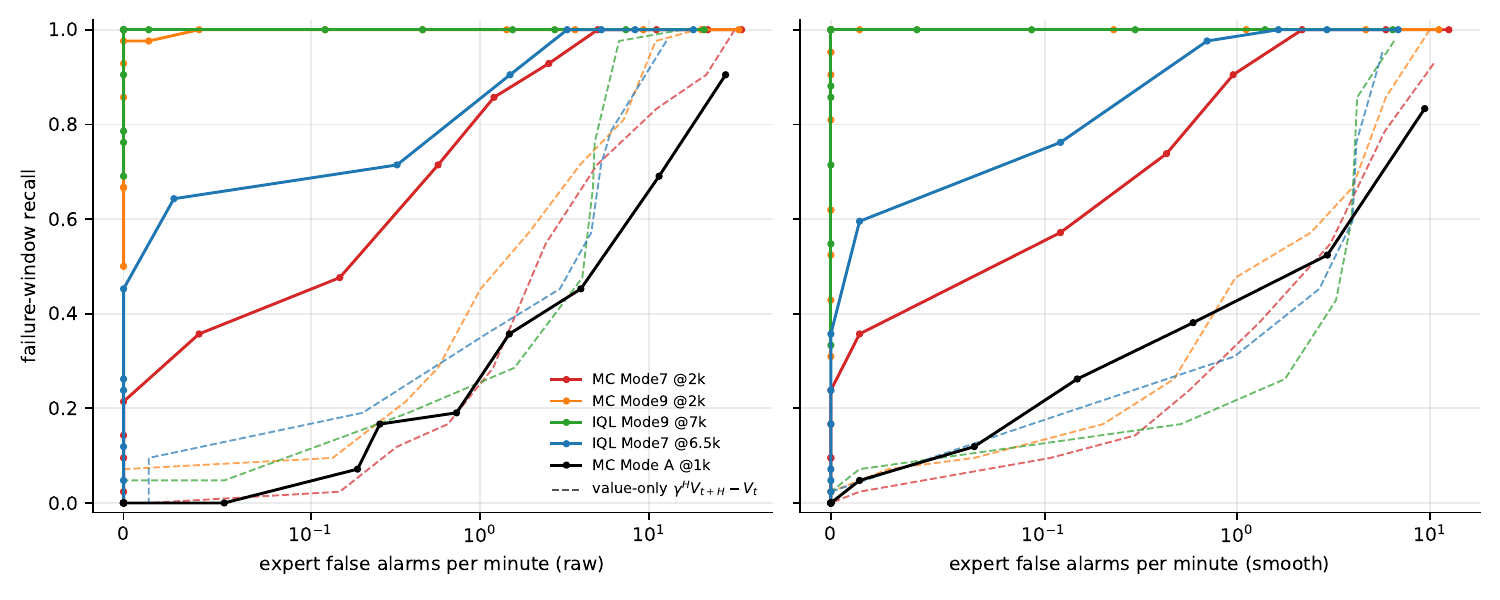}};
\coordinate (ymin) at ($(rocplot.south west)!0.164!(rocplot.north west)$);
\coordinate (ymax) at ($(rocplot.south west)!0.955!(rocplot.north west)$);
\foreach \y in {0.0,0.2,0.4,0.6,0.8,1.0}{
\node[anchor=east,font=\scriptsize,inner sep=1pt] at ($(ymin)!\y!(ymax)+(12pt,0)$) {\y};
}
\node[rotate=90,font=\small] at ([xshift=-12pt]rocplot.west) {Failure-window recall};
\end{tikzpicture}
\caption{Failure-window recall against false alarms per expert minute as the threshold varies for smoothed $d_t$. Solid lines show the full TD residual, and dashed lines show its value-only component.}
\label{fig:realroc}

\medskip

\includegraphics[width=.73\linewidth,trim=0bp 477.4bp 393bp 35bp,clip]{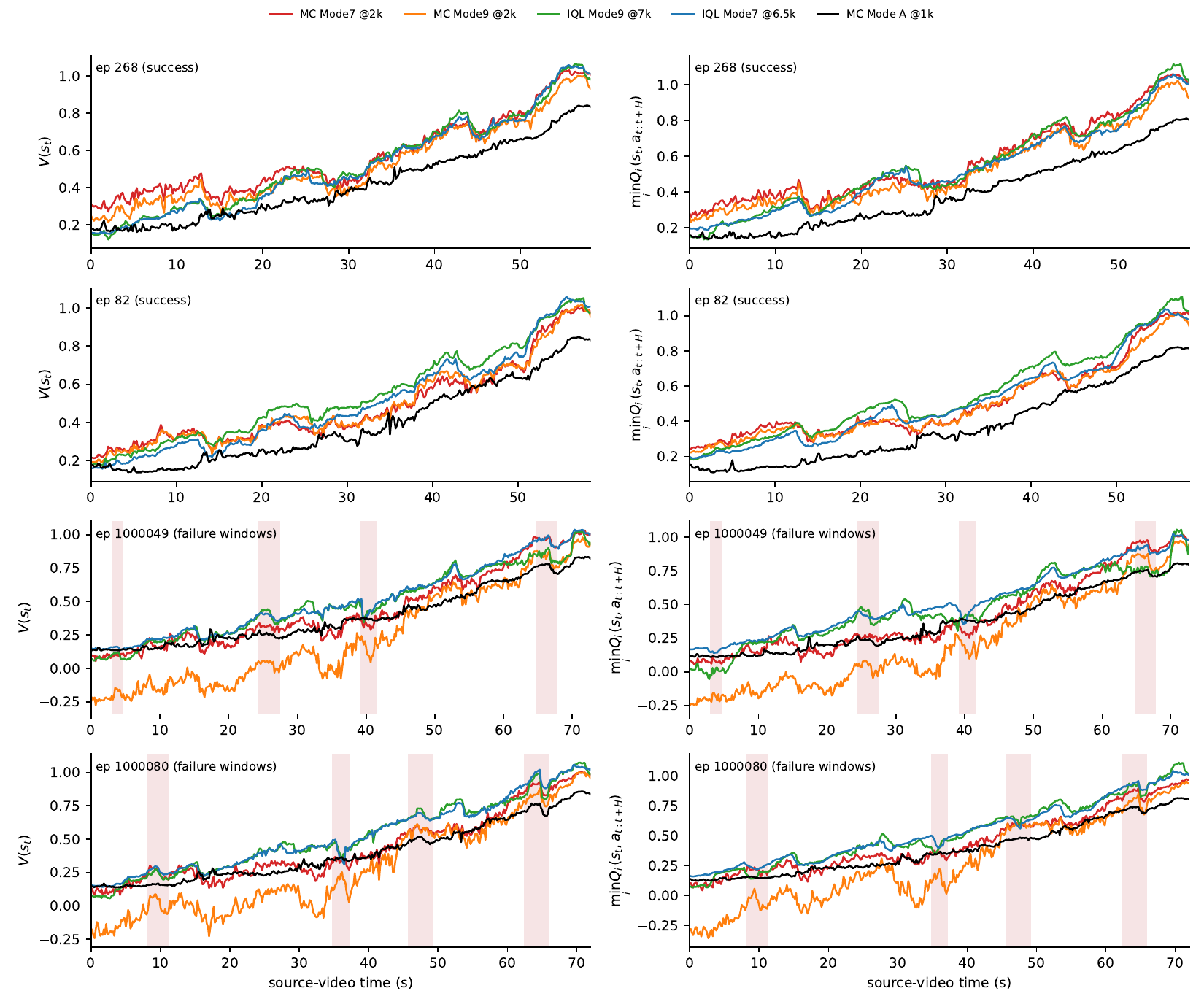}\\[-2pt]
{\small Source-video time (s)}\par\medskip
\includegraphics[width=.73\linewidth,trim=0bp 171.4bp 393bp 338bp,clip]{figs/real_value_q.pdf}\\[-2pt]
{\small Source-video time (s)}
\caption{Value predictions $\hat V_t$ from the five critics on one expert episode (top) and one suboptimal episode (bottom). Shading marks annotated failure windows. Each example is one of the two episodes closest to the median length in its group.}
\label{fig:realvq}
\end{figure}

\runin{Results.}
IQL Mode 7 achieves the highest AUROC ($0.86$) and $76\%$ recall at $0.12$ false alarms per minute. Both Mode 9 critics detect all 42 failure windows, with their residuals including the annotated failure penalty. \Cref{fig:realroc,fig:realvq} show the recall--false-alarm tradeoff and example value trajectories.

\end{document}